\documentclass[10pt,twocolumn,letterpaper]{article}

\usepackage{cvpr}
\usepackage{times}
\usepackage{epsfig}
\usepackage{graphicx}
\usepackage{amsmath}
\usepackage{amssymb}
\usepackage{algorithm}
\usepackage{algpseudocode}
\graphicspath{{./figures/}}
\usepackage{caption} 
\usepackage[pagebackref=true,breaklinks=true,letterpaper=true,colorlinks,bookmarks=false]{hyperref}

\cvprfinalcopy 

\def\cvprPaperID{419} 

\ifcvprfinal\fi
\begin{document}

\title{Stochastic Downsampling for Cost-Adjustable Inference and \\ Improved Regularization in Convolutional Networks}

\author{Jason Kuen\textsuperscript{1}, Xiangfei Kong\textsuperscript{1}, Zhe Lin\textsuperscript{2}, Gang Wang\textsuperscript{3}, Jianxiong Yin\textsuperscript{4}, Simon See\textsuperscript{4}, Yap-Peng Tan\textsuperscript{1}\\\textsuperscript{1}Nanyang Technological University\hspace{25pt}\textsuperscript{2}Adobe Research\hspace{25pt}\textsuperscript{3}Alibaba AI Labs\hspace{25pt}\textsuperscript{4}NVIDIA\\\tt\small \{jkuen001,xfkong,eyptan\}@ntu.edu.sg, zlin@adobe.com, gangwang6@gmail.com,\\\tt\small \{jianxiongy,ssee\}@nvidia.com\vspace*{-15pt}}

\maketitle

\begin{abstract}
   It is desirable to train convolutional networks (CNNs) to run more efficiently during inference. In many cases however, the computational budget that the system has for inference cannot be known beforehand during training, or the inference budget is dependent on the changing real-time resource availability. Thus, it is inadequate to train just inference-efficient CNNs, whose inference costs are not adjustable and cannot adapt to varied inference budgets. We propose a novel approach for cost-adjustable inference in CNNs - Stochastic Downsampling Point (SDPoint). During training, SDPoint applies feature map downsampling to a random point in the layer hierarchy, with a random downsampling ratio. The different stochastic downsampling configurations known as SDPoint instances (of the same model) have computational costs different from each other, while being trained to minimize the same prediction loss. Sharing network parameters across different instances provides significant regularization boost. During inference, one may handpick a SDPoint instance that best fits the inference budget. The effectiveness of SDPoint, as both a cost-adjustable inference approach and a regularizer, is validated through extensive experiments on image classification.
\end{abstract}
\vspace*{-10pt}

\section{Introduction}

Convolutional networks (CNNs) have greatly accelerated the progress of many computer vision areas and applications in recent years. Despite their powerful visual representational capabilities, CNNs are bottlenecked by their immense computational demands. Recent CNN architectures such as Residual Networks (ResNets) \cite{he2016deep,he2016identity} and Inception \cite{szegedy2017inception} require billions of floating-point operations (FLOPs) to perform inference on just one single input image. Furthermore, as the amount of visual data grows, we need increasingly higher-capacity (thus higher complexity) CNNs which have shown to better utilize these large visual data compared to their lower-capacity counterparts \cite{sun2017revisiting}.

There have been works which tackle the efficiency issues of deep CNNs, mainly by lowering numerical precisions (quantization) \cite{hubara2016binarized,rastegari2016xnor,zhu2017trained}, pruning network weights \cite{han2015learning,li2017pruning,yang2017designing,he2017channel,luo2017thinet}, or adopting separable convolutions \cite{jaderberg2014speeding,chollet2017xception,xie2017aggregated}. These methods result in more efficient models which have fixed inference costs (measured in floating-point operations or FLOPs). Models with fixed inference costs cannot work effectively in certain resource-constrained vision systems, where the computational budget that can be allocated to CNN inference depends on the real-time resource availability. When the system is lower in resources, it is preferable to allocate a lower budget for more efficient or cheaper inference, and vice versa. Moreover, in some cases, the exact inference budget cannot be known beforehand during training time.

As a simple solution to such a concern, one could train several CNN models such that each has a different inference cost, and then select the one that matches the given budget at inference time. However, it is extremely time-consuming to train many models, not to mention the computational storage required to store the weights of many models. In this work, we focus on CNNs whose computational costs are dynamically adjustable at inference time. A CNN with cost-adjustable inference only has to be trained once, and it allows users to control the trade-off of inference cost against network accuracy/performance. The different inference instances (each with different inference cost) are all derived from the same model parameters.

\begin{figure*}[t]
	\begin{center}
		\includegraphics[width=0.9\linewidth]{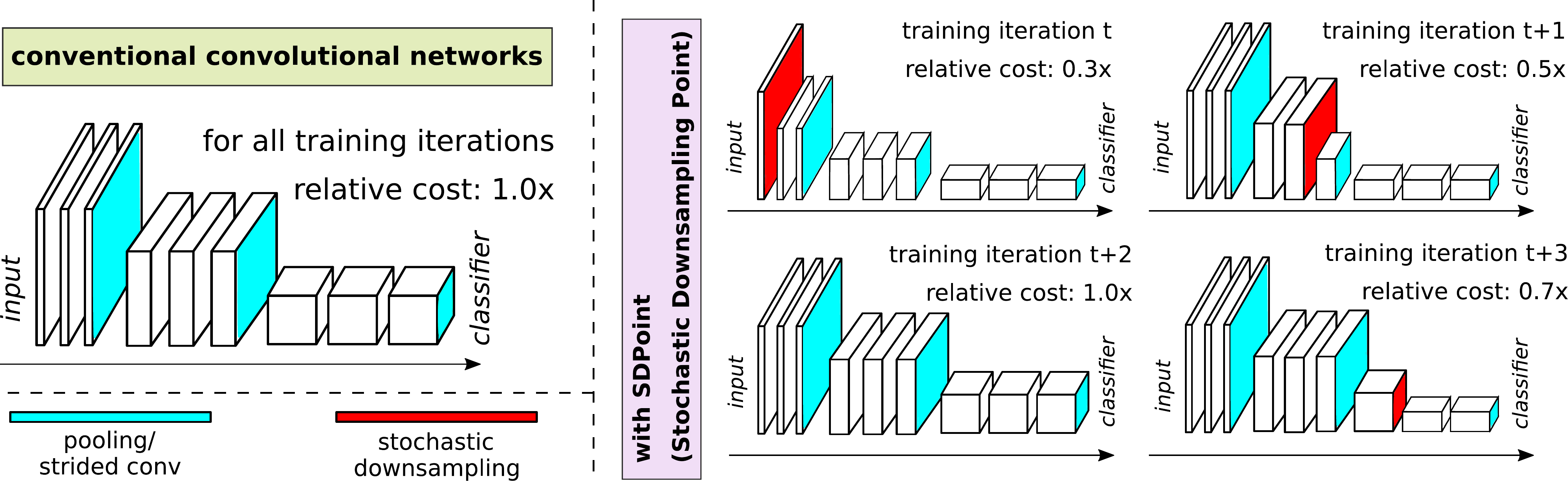}
	\end{center}
	\vspace*{-15pt}
	\caption{Progression of feature map spatial sizes during training of a (\textbf{Left}) conventional CNN, (\textbf{Right}) with SDPoint. The \textit{costs} here refer to computational costs measured in numbers of floating-point operations (FLOPs).}
	\label{fig:SDPoint}
	\vspace*{-10pt}
\end{figure*}

For cost-adjustable inference in CNNs, we propose a novel training method - Stochastic Downsampling Point (SDPoint). A SDPoint instance is a network configuration consisting of a unique downsampling point (layer index) in the network layer hierarchy as well as a unique downsampling ratio. As illustrated in Fig. \ref{fig:SDPoint}, at every training iteration, a SDPoint instance is randomly selected (from a list of instances), and downsampling happens based on the downsampling point and ratio of that instance. The earlier the downsampling happens, the lower the total computational costs will be, given that spatially smaller feature maps are cheaper to process. 

During inference, a SDPoint instance can be deterministically handpicked (among the SDPoint instances seen during training) to match the given inference budget.  Existing approaches \cite{leroux2015resource,teerapittayanon2016branchynet,larsson2017fractalnet} to achieve cost-adjustable inference in CNNs work by evaluating just subparts of the network (e.g., skipping layers or skipping subpaths), and therefore not all network parameters are utilized during cheaper inference. In contrast to existing approaches, SDPoint makes full use of all network parameters regardless of the inference costs, thus making better use of network representational capacity. Moreover, the (scale-related) parameter sharing across the SDPoint instances (each with a different downsampling and downsampling ratio) provides significant improvement in terms of model regularization. On top of these advantages, SDPoint is architecture-neutral, and it adds no parameter or training overheads. We carry out experiments on image classification with a variety of recent network architectures to validate the effectiveness of SDPoint in terms of cost-accuracy performances and regularization benefits. The code to reproduce experiments will be released.


\section{Related Work}\label{relatedwork}

\noindent\textbf{Cost-adjustable Inference:} One representative method to achieve cost-adjustable inference is to train ``intermediate" classifiers \cite{leroux2015resource,lee2015deeply,teerapittayanon2016branchynet} which branch out of intermediate network layers. A lower inference cost can be attained by \textit{early-exiting}, based on the intermediate classifiers' output \textit{confidence} \cite{leroux2015resource} or \textit{entropy} \cite{teerapittayanon2016branchynet} threshold. The lower the threshold is, the lower the inference cost will be, and vice versa. In \cite{leroux2015resource}, intermediate softmax classifiers are trained (second stage) after the base network has been completely trained (first stage). The downside of \cite{leroux2015resource} is that the intermediate classifier losses are not backpropagated for fine-tuning the base network weights. To make the networks more aware of intermediate classifiers, BranchyNet \cite{teerapittayanon2016branchynet} has intermediate classifiers (each with more layers per branch than \cite{leroux2015resource}) and final classifier trained jointly, using a weighted sum of classification losses. Unlike these works, our SDPoint method relies on the same final classifier for different inference costs. FractalNets \cite{larsson2017fractalnet} which are CNNs designed to have many parallel subnetworks or ``paths" which can be stochastically dropped for regularization during training. For cost-adjustable inference, some FractalNet's ``paths'' can be left out. But the path-dropping regularization gives inconsistent/marginal improvements if data augmentation is being used.

Another line of work somehow related to cost-adjustable inference is \textit{adaptive computation} in recurrent networks \cite{graves2016adaptive} and CNNs \cite{figurnov2017spatially}. The inference costs of adaptive computation networks are adaptive to the given inputs - harder examples cost more than easier ones. The learned policies of choosing the amount of computation however cannot be modified during inference for cost-adjustable inference.\\
 
\noindent\textbf{Stochastic Regularization:} Our work is closely related to stochastic regularization methods which apply certain stochastic operations to network training for regularization. Dropout \cite{srivastava2014dropout} drops network activations, while DropConnect \cite{wan2013regularization} drops network weights. Stochastic Depth \cite{huang2016deep} allows nonlinear residual building blocks to be dropped during training. These 3 methods are similar in the way that during inference, all stochastically dropped elements (activations, weight, residual blocks) are to be present. For any of the methods, its different stochastic instances seen during training have rather comparable forward pass costs, making them unfit for cost-adjustable inference.\\

\noindent\textbf{Multiscale parameter-sharing:} Multiscale training of CNNs, first introduced by \cite{he2014spatial} is quite similar to SDPoint. In the training algorithm of \cite{he2014spatial}, the network is trained with $224\times224$ and $180\times180$ images alternatively (one scale per epoch). The same idea has also been applied to CNN training for other tasks \cite{chen2017deeplab,redmon2017yolo9000}. While multiscale training downsamples the input images to different sizes, SDPoint only downsamples feature maps (at feature level). Downsampling at feature level encourages earlier network layers to learn to better preserve information, to compensate for loss of spatial information caused by stochastic downsampling later. This does not apply to multiscale training, where the input images are downsampled through interpolation operations which happen before network training takes place.

\section{Preliminaries: Conventional CNNs with Fixed Downsampling Points}\label{prelim}

Conventionally, downsampling of feature maps happens in CNNs at several predefined fixed locations/points in the layer hierarchy, depending on the architectural designs. For example, in ResNet-50, spatial pooling (happens after the first ReLU layer, and after the last residual block) and strided convolutions (or convolution with strides $>1$ which happens right after the 3rd, 7th, and 13th residual blocks) are used to achieve downsampling. Between these downsampling layers are \textit{network stages}. Downsampling in CNNs trades low-level spatial information for richer high-level semantic information (needed for high-level visual tasks such as image classification) in a gradual fashion.

During network inference, these fixed downsampling points have to be followed exactly as how they are configured during training, for optimal accuracy performance. In this work, we go beyond fixed downsampling points - we develop a novel stochastic downsampling method named Stochastic Downsampling Point (SDPoint) which does not restrict downsampling to happen every time at same fixed points in the layer hierarchy. The proposed method is complementary to the fixed downsampling points in existing network architectures, and do not replace them. SDPoint can be simply plugged into existing network architectures, and no major architectural modifications are required. 

\section{Stochastic Downsampling Point}\label{sdp}

A Stochastic Downsampling Point (SDPoint) instance has a unique downsampling point $ p \in \mathbb{Z}$ and a unique downsampling ratio $ r \in \mathbb{R}$ which are stochastically/randomly selected during network training. A $p$  and a $r$ are stochastically selected at the beginning of each network training iteration, and downsampling occurs to the selected point (based on the selected ratio) for all samples in the current training mini-batch. The downsampling points and a downsampling ratios will be discussed more thoroughly in the upcoming sections. Downsampling is performed by a downsampling function $ D(\cdot) $ which makes use of some downsampling operations. When the selected point falls at the lower layer in the layer hierarchy, the downsampling happens earlier (in the forward propagation), causing quicker loss of spatial information in the feature maps, but more computation savings. Conversely, spatial information can be better preserved at higher computational costs, if the stochastic downsampling happens later. 

SDPoint can effectively turn the feature map spatial sizes right before prediction layers to be different from original sizes, and this could cause shape incompatibility between the prediction layer weights (as well as labels) and the convolutional outputs (before prediction layers). To prevent this, we preserve the feature map spatial size in the last network stage, regardless of stochastic downsampling taking place or not, by adjusting convolution strides and/or pooling sizes accordingly. For example, in image classification networks, we consider the global average pooling layer \cite{lin2014network} and the final classification layer to be the last network stage. Therefore, regardless of the spatial size (variable due to SDPoint) of the incoming feature maps, we globally pool them to have spatial size of $ 1\times1 $.

\subsection{Downsampling Operation}\label{downop}
As discussed in Sect. \ref{prelim}, the downsampling operation employed in $D(\cdot)$ can be either pooling \cite{boureau2010theoretical} (\textit{average} or \textit{max} variations) or strided convolution. We opt for average pooling (the corresponding downsampling function is denoted as $D_{\text{avg}}(\cdot)$), rather than strided convolutions or max pooling for several reasons. Strided convolutions are the preferred way to do downsampling in recent network architectures, because they add extra parameters (convolution weights) and therefore improving the representational capability. In this work, we want to rule out the possible performance improvements from increase in parameter numbers (rather than the SDPoint itself). Moreover, strided convolutions with integer-valued strides cannot work well with arbitrary downsampling ratios (see Sect. \ref{ratios}). On the other hand, average pooling is preferred over max pooling in this paper due to the fact that max pooling itself is a form of non-linearity. Using max pooling as the downsampling operation could either push for a greater non-linearity in the network (positive outcome) which is unfair to the baselines, or could exacerbate the vanishing gradient problem \cite{hochreiter1991untersuchungen} commonly associated with deep networks (negative outcome). Besides, the effectiveness of average pooling has been validated through its extensive roles in recent CNN architectures (e.g., global average pooling \cite{lin2014network,he2016deep}, DenseNets' transition \cite{huang2017densely}).

\subsection{Downsampling Points}\label{pointcandidates}
At every training iteration, a downsampling point $p$ for a SDPoint instance can be drawn from a discrete uniform distribution on a set of predefined downsampling point indices $\mathit{P} = \{0,1,2,...,\mathcal{N}\text{-}1,\mathcal{N}\}$, with $ \mathcal{N}+1 $ number of points. In this work, the downsampling point candidates are the points between two consecutive CNN ``basic building blocks", mirroring the placements of \textit{fixed downsampling} layers in conventional CNNs. We keep the original network (without stochastic downsampling) as an instance by assigning the index $p=0$ to it, so that we can perform full-cost inference later. Let $ F(\cdot) $ denote the function carried out by the $i$-th basic building block, $ \mathbf{w}_i $ denote the network weights involved in the block. For a given input $ \mathbf{x}_i $ and downsampling ratio $ r $, the downsampling is carried out as following:

\begin{equation}
\mathbf{y}_i = D_{\text{avg}}(F(\mathbf{x}_i; \mathbf{w}_i); s_i, r)
\end{equation}
to obtain the output $ \mathbf{y}_i  $.  The downsampling switch denoted as $ s_i \in\{\texttt{True},\texttt{False}\} $ is turned on if $p=i$.

For non-residual CNNs (e.g., VGG-Net \cite{simonyan2015very}), the basic building block comprises 3 consecutive \textit{convolutional}, \textit{Batch Normalization} (BN) \cite{ioffe2015batch}, \textit{non-linear activation} layers. On the other hand, for residual networks, residual blocks are considered as the basic building blocks. the downsampling point $ p $ can be stochastically selected to be any point between any 2 basic building blocks in the network, where downsampling happens. Since a residual block involves two streams of information - (i.) the identity skip connection and (ii.) the non-linear function consisting of several network layers, we apply stochastic downsampling function $ D_{\text{avg}}(\cdot) $ to the point right after the residual addition operation. We also experiment with Densely Connected Networks (DenseNets) \cite{huang2017densely} in this paper. For DenseNets, the SDPoint downsampling points are the points right behind each block concatenation operation, mirroring the \textit{fixed downsampling} in DenseNets.

In principle, each mini-batch sample could have its unique downsampling point $p_i$ (for stronger stochasticity), but due to practical reasons (e.g., training efficiency, ease of implementation), we resort to using the same $p_i$ for all samples in a mini-batch. While it is possible to have more than one downsampling points in each training iteration, the number of possible combinations or SDPoint instances would become excessively large. Some of the instances would deviate too much from the original network, in terms of computational cost and accuracy performance. We opt for single \textbf{stochastic downsampling point} in this work.

\subsection{Downsampling Ratios}\label{ratios}
We consider a set of downsampling ratios $ \mathit{R} $, which the SDPoint instance can stochastically draw a downsampling ratio $ r $ from, for use at current training iteration. As with Sect. \ref{pointcandidates}, downsampling ratios are drawn according to discrete uniform distributions. The ratios cannot be too low that they hamper the training convergence (due to parameter-sharing unfeasibility). And, we consider only a small number of downsampling ratios in $ \mathit{R} $ to prevent an excessive number of SDPoint instances, which would cause great difficulty in experimentally evaluating all SDPoint instances for cost-adjustable inference. A recent experimental study \cite{mishkin2017systematic} on CNNs finds that it is sufficient to make qualitative conclusions about optimal network structure that hold for the full-sized ($ 224\times224$ image resolution) ImageNet \cite{russakovsky2015imagenet} classification task, by using just $ 128\times128 $ (roughly half the original resolution) input images. Conceivably, the same network structure/architecture that works well with a certain image resolution is likely to work well with a resolution double/half of that. Motivated by the above-mentioned heuristics and experimental finding, we come up with the downsampling ratio set $ \mathit{R} = \{0.5, 0.75\}$. The same ratios have also been used by \cite{chen2017deeplab} for ``multiscale-input" semantic segmentation. The same hyperpameter $ \mathit{R} $ is used across all experiments in this paper.

Downsampling with such fractional downsampling ratios cannot be trivially achieved with integer-valued pooling hyperparameters. For example, pooling a $28\times28$ feature map to a $21\times21$ one (with $ r $ of $0.75$ and minimal overlaps) cannot be easily done by tuning just the pooling size and stride. To this end, we adopt a spatial pooling strategy (which works along with the pooling choice in Sect. \ref{downop}) akin to that of Spatial Pyramid Pooling \cite{he2014spatial} that generates fixed-length representation via adaptive calculations of pooling sizes and strides.

\begin{algorithm}
	\caption{: Training with SDPoint}
	\label{trainingwithSDP}
	\begin{algorithmic}[1] %
		\State $ \mathit{P} = \{0,1,2,...,\mathcal{N}\text{-}1,\mathcal{N}\} $\Comment{Downsampling Points}
		\State $ \mathit{R} = \{0.5, 0.75\} $\Comment{Downsampling Ratios}
		\While {given a training mini-batch $ \mathbf{x}$}
		\State {Randomly draw $ p $ from $P$}
		\State {Randomly draw $ r $ from $R$}
		\State $ \mathbf{x}_{1} = \mathbf{x} $
		\For {$ i \in \{1,2,...,\mathcal{N}\text{-}1,\mathcal{N}\}$} \Comment{Forward pass}
		\If{$i = p$}
		\State {$s_i = $ \texttt{True}}
		\Else
		\State {$s_i = $ \texttt{False}}
		\EndIf
		\State {$ \mathbf{x}_{i+1} = D_{\text{avg}}(F(\mathbf{x}_i; \mathbf{w}_i); s_i, r) $}
		\EndFor
		\State {Compute \textit{loss} with $ \mathbf{x}_{\mathcal{N}+1} $}
		\State {Backward pass}
		\State {Parameter updates}
		\EndWhile\label{euclidendwhile}
	\end{algorithmic}
\end{algorithm}

\vspace*{-8pt}
\subsection{Training with SDPoint}
SDPoint gives rise to a new training algorithm for CNNs. The training algorithm consolidating all the previously introduced SDPoint concepts is given in Algorithm \ref{trainingwithSDP}. $ F (\cdot) $ denotes the generic nonlinear building network block in CNNs. For simplicity sake, we omit the other network layers which are not basic building blocks - typically the starting and ending layers. In a nutshell, Algorithm \ref{trainingwithSDP} shows that whenever a building block index $ i $ is equal to the downsampling point $ p $, the downsampling switch $ s $ is turned on. Stochastic downsampling then happens to the output of $ i $-th building block, with the stochastic downsampling ratio $ r $. It is important to point out that the (stochastic) downsampling does not happen, if $ p $ is drawn to be $ 0 $, allowing the network to work in its original ``unadulterated" form.

\subsection{Regularization}\label{regularization}
SDPoint can be seen as a regularizer for CNNs. When stochastic downsampling takes place, the receptive field size becomes larger and it causes a sudden shrinkage of spatial information in the feature maps. The network has to learn to adapt to such variations during training, and perform \textbf{parameter-sharing} across the downsampled feature maps and the originally sized feature maps (when $ p = 0 $). In addition to robustness in terms of receptive field size and spatial shrinkage, SDPoint also necessitates the convolutional layers to accommodate for different ``\textit{padded pixel} to \textit{non-padded pixel}" ratios. 
For example, applying a $ 3 \times 3 $ convolutional filter (with zero-padding of 1) to a $ 8 \times 8 $ feature map gives a padded-pixel ratio of 0.44, compared to 0.56 ratio resulted from applying the same filter to $ 6 \times 6 $ feature map. Zero-padded pixels are quite similar to the \textit{zero-ed out} activations caused by Dropout \cite{srivastava2014dropout}, in the sense that they both are missing values. Thus, a higher padded-pixel ratio is akin to having a higher number of \textit{dropped-out} activations, vice versa. This form of variation provides further regularization boost. Experimentally, we find that even with the use of heavy data augmentation - such as ``scale + aspect ratio" augmentation \cite{szegedy2015going,szegedy2017inception}, SDPoint can still help.

\section{Cost-adjustable Inference}
A network that can perform inference at different computational costs depending on the user requirements, is considered to be capable of \textit{cost-adjustable inference}. Opting for a lower inference cost usually results in a lower prediction accuracy, and vice versa. SDPoint naturally supports cost-adjustable inference, given that SDPoint instances have varying computational costs, given the different downsampling point locations and downsampling ratios. More importantly, the instances have all been trained to minimize the same prediction loss, and this helps them to work relatively well for inference. During inference, one may handpick a SDPoint instance (with its downsampling point $ p $ and downsampling ratio $ r $) to make the inference cost fit a particular inference budget.\\

\noindent\textbf{5.1\hspace{5pt}Instance-Specific Batch Normalization} As mentioned in Sect. \ref{sdp}, SDPoint instances are trained in such a way that every training mini-batch and iteration shares the same SDPoint instance. For a SDPoint instance, the prediction and loss minimization during training are based on the Batch Normalization (BN) statistics (means and standard deviations) of that particular instance. Therefore, using the BN statistics accumulated over many training iterations (and thus many different SDPoint instances) for inference causes inference-training ``mismatch". A similar form of inference-training ``mismatch" caused by BN statistics has also been observed by \cite{singh2016swapout} in another context. The BN statistics required for one SDPoint instance should differ from that of another instance.  When using the same (accumulated) BN statistics to perform cost-adjustable inference, the inference accuracies could be jeopardized. 

To address the ``mismatch" issue, we compute SDPoint instance-specific BN statistics, and use them for cost-adjustable inference. Disentangling the different SDPoint instances by unsharing BN statistics makes the inference more accurate. The computational storage overhead resulted from \textit{instance-specific BN} statistics is relatively low, as BN statistics of some earlier layers can be shared\footnote{refer to supplementary materials for more about storage overheads.} among certain SDPoint instances that downsample at later layers. 

\section{Experiments}
Experiments are carried out on image classification tasks to evaluate SDPoint. We consider image classification datasets with varying dataset scales in terms of numbers of categories/classes and sample counts: CIFAR-10 \cite{krizhevsky2009learning} (50k training images, 10k validation images, 10 classes), CIFAR-100 \cite{krizhevsky2009learning} (50k training images, 10k validation images, 100 classes), ImageNet \cite{russakovsky2015imagenet} (1.2M training images, 50k validation images, 1000 classes). For inference cost comparison, we measure the model costs in terms of floating-point operation numbers (FLOPs) needed for forward propagation of single image. We treat \textit{addition} and \textit{multiplication} as 2 separate operations. Implementations are in PyTorch \cite{pytorch}.

\begin{figure*}[t]
	\begin{center}
		\includegraphics[width=0.9\linewidth]{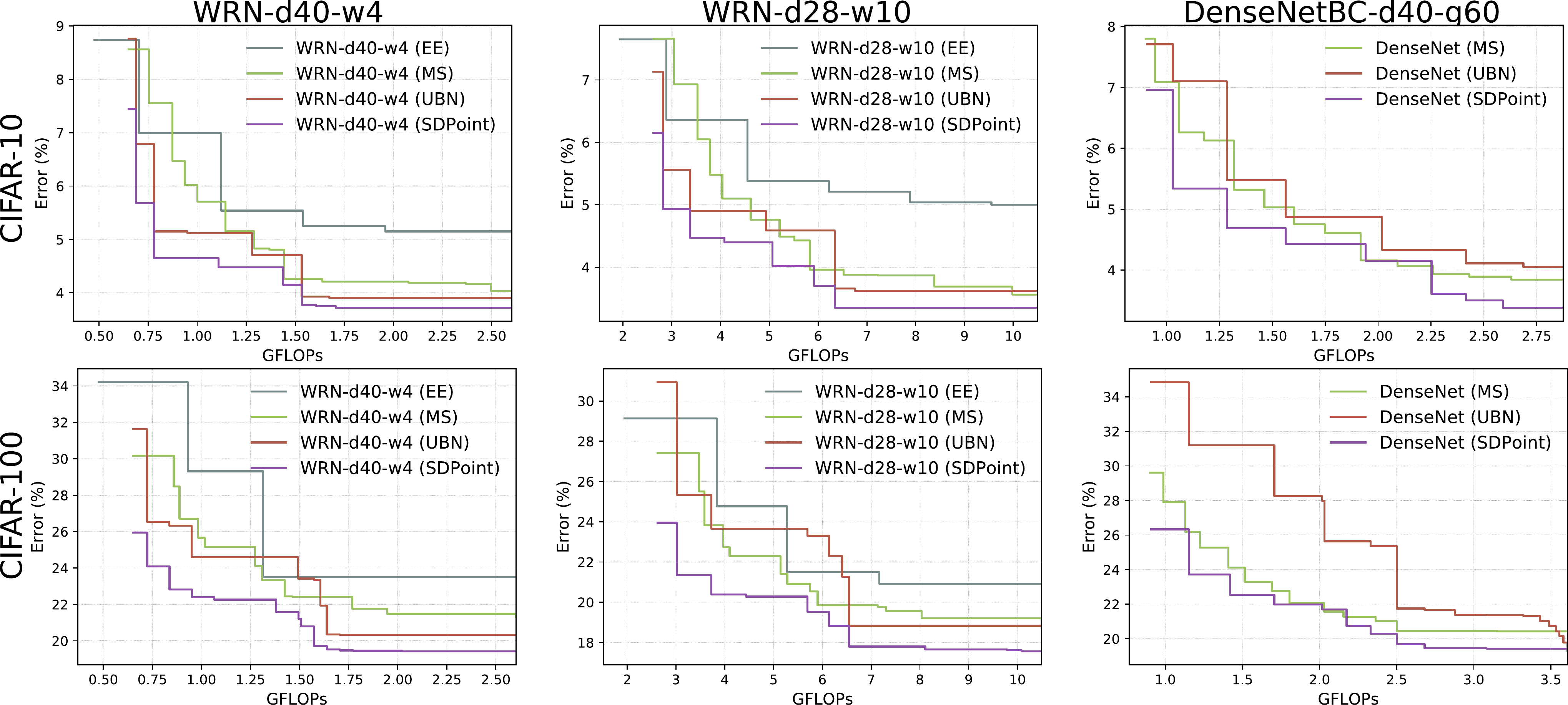}
	\end{center}
	\vspace*{-15pt}
	\caption{WRNs' and DenseNetBC's cost-error plots on CIFAR-10 (\textbf{Top}) and CIFAR-100 (\textbf{Bottom}). It is observed that models trained with SDPoint consistently outperform their non-SDPoint counterparts, given the same computational budgets.}
	\label{fig:cifar}
	\vspace*{-12pt}
\end{figure*}

\subsection{CIFAR}\label{cifar}
For CIFAR-10 and CIFAR-100, the baseline architectures are Wide-ResNet \cite{zagoruyko2016wide} (WRN-d28-w10 and WRN-d40-w4) and DenseNetBC-d40-g60 \cite{huang2017densely}. `d', `w', `g' stand for the network depth, widen factor of WRN, and growth rate of DenseNetBC, respectively. The training hyperparameters (e.g., learning rates, schedules, batch sizes, augmentation) follow the ones in original papers, except for training epoch numbers which we fix to 400 for all. The original learning rate schedules still apply (e.g., learning rates are dropped at 50\% and 75\% of total number of training epochs). The numbers of SDPoint downsampling points ($\mathcal{N}$) for \{WRN-d28-w10, WRN-d40-w4, and DenseNetBC-d40-g60\} are \{12, 18 ,12\} respectively. As mentioned in Sect. \ref{ratios}, the downsampling ratios are drawn uniformly from $ \mathit{R} = \{0.5, 0.75\} $.\\ 

\begin{table}[t]
	\centering
	\resizebox{1.025\linewidth}{!}{
		\begin{tabular}{| r | r | r | r | r | r |} 
			\hline
			Model & \# Params & GFLOPs & CIFAR-10 & CIFAR-100 \\
			\hline\hline
			ResNeXt-d29-c08 \cite{xie2017aggregated} & 34.4M & 10.8 & 3.65 & 17.77 \\ 
			\hline
			ResNeXt-d29-c16 \cite{xie2017aggregated} & 68.1M & 21.4 & 3.58 & 17.31 \\
			\hline
			DenseNetBC-d250-g24 \cite{huang2017densely} & 15.3M & 10.1 & 3.62 & 17.60 \\
			\hline
			DenseNetBC-d190-g40  \cite{huang2017densely} & 25.6M & 18.7 & 3.46 & \textbf{17.18} \\
			\hline\hline
			WRN-d40-w4 \cite{zagoruyko2016wide} & 8.9M & 2.6 & 4.29 & 20.78 \\
			\hline
			WRN-d40-w4 \cite{zagoruyko2016wide} & 8.9M  & 2.5/2.6 & 3.73  & 19.55\\
			with \textbf{SDPoint} &  & & ($\downarrow$ 0.56)  & ($\downarrow$ 1.23)\\
			\hline
			WRN-d28-w10 \cite{zagoruyko2016wide} & 36.5M & 10.5 & 3.84 & 18.51 \\
			\hline
			WRN-d28-w10 \cite{zagoruyko2016wide} & 36.5M & 10.5 & 3.86 & 18.05 \\
			with Dropout \cite{srivastava2014dropout} & &  & ($\uparrow$ 0.02) & ($\downarrow$ 0.46)\\
			\hline
			WRN-d28-w10 \cite{zagoruyko2016wide} & 36.5M & 6.5/10.1 & \textbf{3.35} & 17.53\\
			with \textbf{SDPoint} & &  & ($\downarrow$ 0.49) & ($\downarrow$ 0.98)\\
			\hline
			DenseNetBC-d40-g60 \cite{huang2017densely} & 4.3M & 3.6 & 3.99 & 20.00 \\
			\hline
			DenseNetBC-d40-g60 \cite{huang2017densely} & 4.3M & 2.7/3.6 & 3.39 &19.25\\
			with \textbf{SDPoint} & & & ($\downarrow$ 0.60)  & ($\downarrow$ 0.75)\\
			\hline
		\end{tabular}
	}
	\caption{CIFAR-10 and CIFAR-100 validation errors (\%). The GFLOPs with 2 values separated by ``/" are for CIFAR-10 and CIFAR-100 respectively.}
	\vspace*{-13pt}
	\label{table:cifar}
\end{table}

\noindent\textbf{6.1.1\hspace{5pt}Baseline Comparison:}
We compare SDPoint with some baseline methods related to ours, in terms of cost-adjustable inference performance. The classification error-cost performance plots on CIFAR-10 and CIFAR-100 are shown in Fig. \ref{fig:cifar}. Note that for SDPoint and baseline methods, not all instances of the same model appear on the plots; if a higher-cost instance performs worse than any lower-cost instance, it is not shown. Each model (evaluated on a dataset) is trained only once to obtain its cost-error plot.\\

\noindent\textbf{(i) Early-Exits (EE)}  We train models based on the WRN with intermediate classifiers (branches) which allow early-exits (EE), following the design of BranchyNet \cite{teerapittayanon2016branchynet}. Each \textit{network stage} in the main network has two evenly spaced branches, and the branches each have single-repetition of building block per \textit{branch network stage}. The blocks in the branches follow the same hyperparameters (e.g., \#channels) as the blocks in the original network. For cost-adjustable inference, we evaluate every branch, and make all samples ``exit" at the same branch. The early-exit models have considerably more parameters than both the baseline models and SDPoint-based models. We conjecture that the relatively worse performance of EE is due to lack of full network parameter ultilization. Also, EE forces CNN features to be classification-ready in early stage, thus causing higher layers to rely heavily on the classification-ready features, instead of learning better features on their own.  \\

\noindent\textbf{(ii) Multiscale Training (MS)} Multiscale (MS) training is a baseline method inspired by \cite{he2014spatial, chen2017deeplab, redmon2017yolo9000}. The input images are downsampled using bilinear interpolations, to an integer-valued size randomly chosen from sizes ranging from half ($ 16 \times 16$) to full size ($ 32 \times 32$), with step size of 1 pixel. This is done for every training iteration, similar to SDPoint. The number of ``instances" (16) resulted from multiscale training is close to the downsampling point numbers of applying SDPoint to WRNs and DenseNetBC(s). Also, the ranges of cost-adjustable inference costs among them are comparable. Instance-specific BN statistics are applied. The cost-adjustable performance of MS consistently trails behind that of SDPoint, as input downsampling causes more drastic information loss than feature map downsampling (see Sect. \ref{relatedwork}).\\

\noindent\textbf{(iii) Uniform Batch Normalization (UBN)} To validate the effectiveness of SDPoint instance-specific BN, we show the results of a SDPoint baseline whose BN statistics are averaged from many training iterations, and are uniform for all of its instances. There are consistent classification performance gaps between using UBN statistics and instance-specific BN statistics, suggesting that it is preferable to keep instance-specific statistics for inference. \\

\noindent\textbf{6.1.2\hspace{5pt}State-of-the-art Comparison:} Table \ref{table:cifar} reports the CIFAR validation results of state-of-the-art (SOTA) ResNeXt \cite{xie2017aggregated} and DenseNetBC \cite{huang2017densely} models, for comparison with ours. For each SDPoint-enabled model, we show the results (giga-FLOPs, classification errors) from the best-performing SDPoint instance among its instances. Notably, WRN-d28-w10 with SDPoint is competitive to SOTA models on CIFAR-100, and it outperforms them on CIFAR-10. Overall, SDPoint considerably improves classification performance without bringing in additional parameters and computational costs, unlike the SOTA models which require about $2\times$ model complexity to attain slight improvements. In fact, the best SDPoint-enabled models on CIFAR-10 have reduced inference costs (FLOPs). We reckon that a prolonged preservation of spatial details (i.e., no early downsampling) in CNN feature maps is not crucial to a dataset with relatively low label complexity such as CIFAR-10. This reveals a drawback of current practice of using CNNs in ``one-size-fits-all" fashion.

\begin{table}[t]
	\centering
	\resizebox{1.025\linewidth}{!}{
		\begin{tabular}{| r | r | r | r | r | r |}
			\hline
			Model  & \# Params & GFLOPs & Top-1 & Top-5 \\
			\hline\hline
			ResNeXt-d101-c64 \cite{xie2017aggregated} & $\sim$89M & $\sim$32 & \textbf{20.4} & \textbf{5.3} \\ 
			\hline
			DenseNetBC-d264 \cite{pleiss2017memory} & $\sim$73M & $\sim$26 & \textbf{20.4} & - \\ 
			\hline\hline
			ResNeXt-d101-c32 \cite{xie2017aggregated} & 44.3M & 16.0 & 21.2 & 5.6 \\ 
			\hline
			ResNeXt-d101-c32 \cite{xie2017aggregated}  & 44.3M & 16.0  & \textbf{20.4} & \textbf{5.3}  \\ 
			with \textbf{SDPoint} & & & ($\downarrow$ 0.8) &  ($\downarrow$ 0.3) \\
			\hline
			PreResNet-d101 \cite{he2016identity} & 44.7M & 15.7  & 22.0 & 6.1 \\	
			\hline
			PreResNet-d101 \cite{he2016identity}  & 44.7M & 15.7 & 21.4 & 5.6  \\
			with \textbf{SDPoint}  & & & ($\downarrow$ 0.6) & ($\downarrow$ 0.5)  \\
			\hline\hline
			PreResNet-d101 \cite{he2016identity} & 45.0M & 11.1 & 24.4 & 7.2 \\	
			with SACT \cite{figurnov2017spatially} & & & &  \\
			\hline
			PreResNet-d101 \cite{he2016identity}  & 44.7M & 7.7 & 24.3 & 7.2 \\	
			with \textbf{SDPoint} & & & &  \\
			\hline
		\end{tabular}
	}
	\caption{ImageNet top-1 and top-5 validation errors (\%), with model parameter numbers and giga-FLOPs (GFLOPs).}
	\vspace*{-13pt}
	\label{table:imagenet}
\end{table}

\subsection{ImageNet}
\begin{figure}[t]
	\begin{center}
		\includegraphics[width=0.9\linewidth]{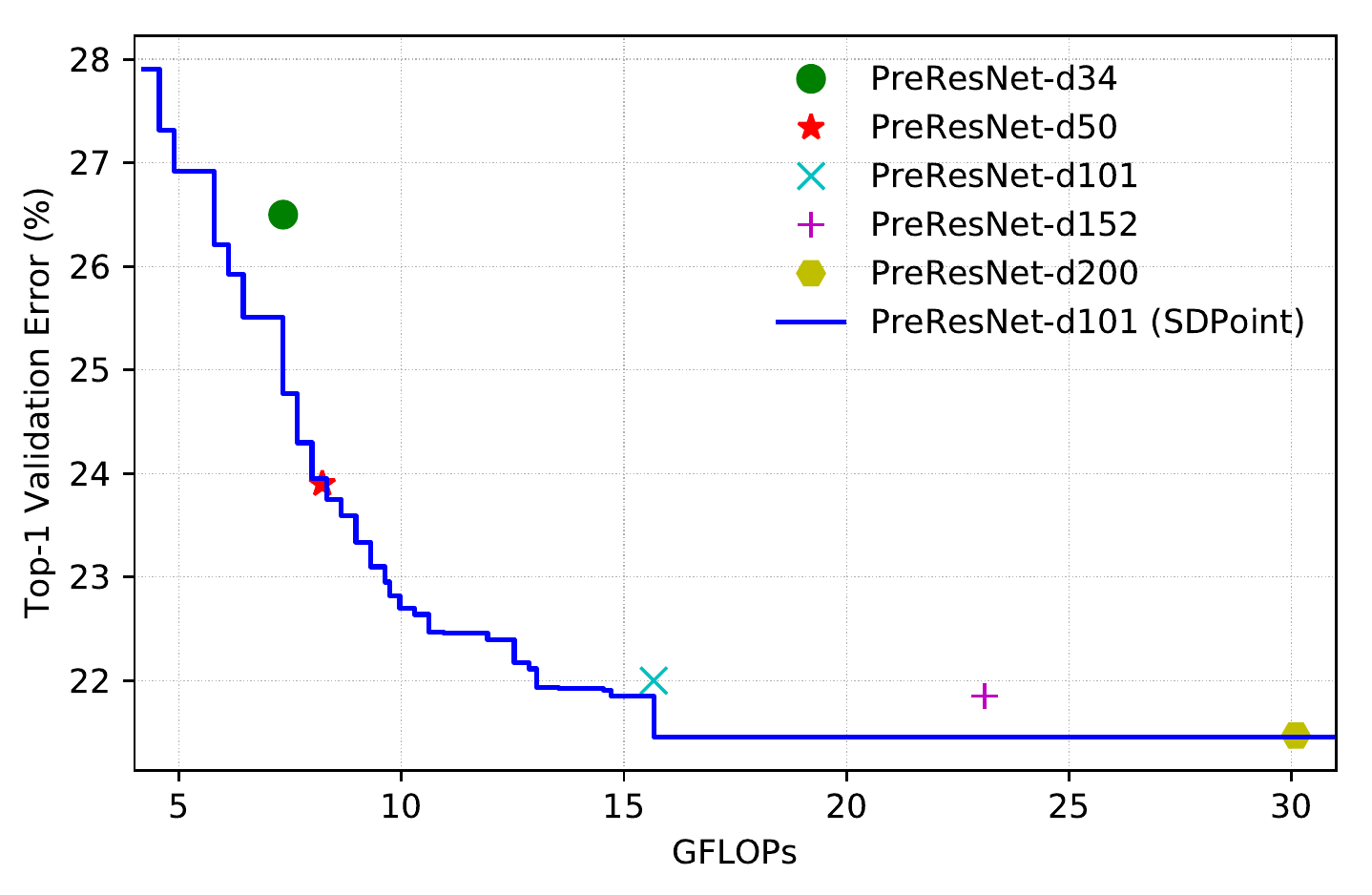}
	\end{center}
		\vspace*{-15pt}
	\caption{PreResNets' \cite{he2016identity} cost-error plots on ImageNet. PreResNet-d101 (SDPoint) only has to be trained once (as a single model), while the baseline models (without SDPoint) has to be trained separately with huge training and storage costs.}
	\label{fig:preresnet}
	\vspace*{-16pt}
\end{figure}

\begin{figure}[t]
	\begin{center}
		\includegraphics[width=0.9\linewidth]{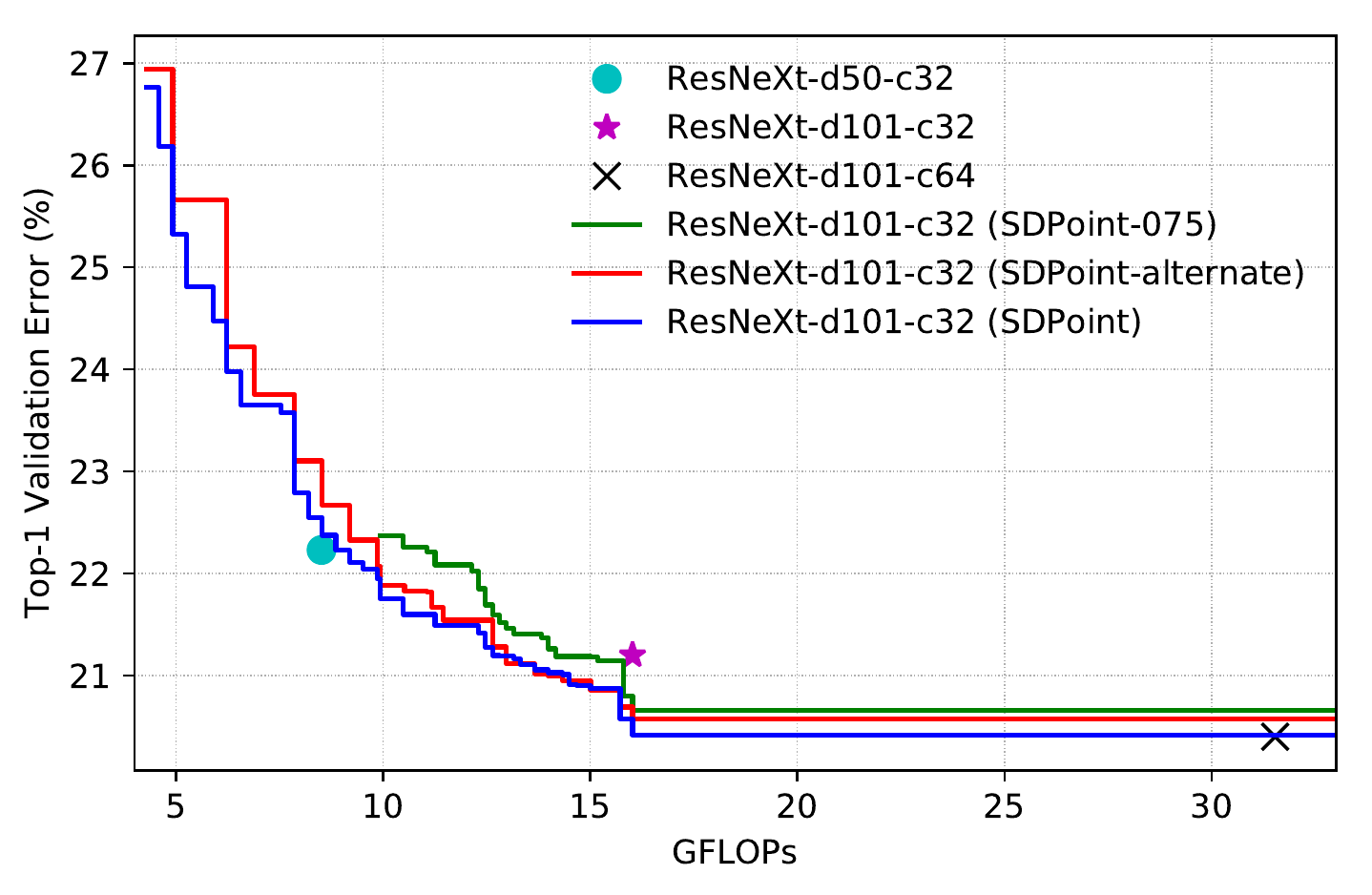}
	\end{center}
	\vspace*{-15pt}
	\caption{ResNeXts' \cite{xie2017aggregated} cost-error plots on ImageNet. Like Fig. \ref{fig:preresnet}, any of ResNeXt-d101-c32 (SDPoint..) only has to be trained once (as a single model).}
	\label{fig:resnext}
	\vspace*{-13pt}
\end{figure}

We consider ResNeXt-d101-c32 \cite{xie2017aggregated} and PreResNet-d101 \cite{he2016identity} as baseline architectures. `c' stands for ResNeXt's cardinality. With SDPoint, there are 33 downsampling points ($\mathcal{N}$) per model. We train the models on ImageNet-1k \cite{russakovsky2015imagenet} training set, and evaluate them on the validation set (224$\times$224 center crops). All models are trained using training hyperparameters and ``scale + aspect ratio" augmentation \cite{szegedy2015going} identical to \cite{xie2017aggregated}. Note that we do not allocate more training epochs to models with SDPoint. The cost-error plots are given in Fig. \ref{fig:preresnet} and \ref{fig:resnext}, for PreResNet-d101 and ResNeXt-d101-c32 respectively, along with some \textbf{fixed-cost} \& \textbf{carefully designed}\footnote{model hyperparameters are carefully chosen by the authors \cite{he2016identity,xie2017aggregated} to optimize accuracy performances under some budget constraints.} baseline models from the same architecture families. Overall, models trained with SDPoint can roughly match the performance of baseline models in the lower-cost range, and surpass them in the upper-cost range. Notably, to obtain cost-error plots, SDPoint-enabled models only have to be trained once. The baseline models are \textbf{trained separately}, resulting in a huge total number of epochs (\#models $\times$ \#epochs per model) and storage cost.\\

\begin{figure*}[t]
	\begin{center}
		\includegraphics[width=0.87\linewidth]{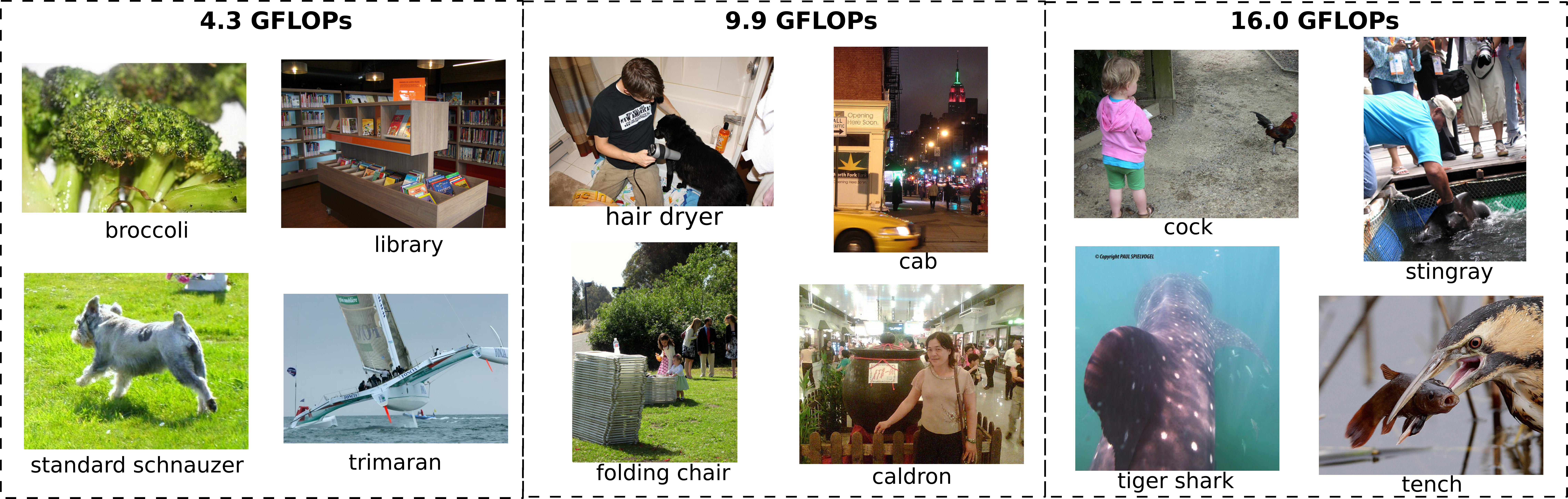}
	\end{center}
	\vspace*{-15pt}
	\caption{Some Imagenet validation examples grouped according to the \textbf{minimum} inference costs (FLOPs) required by ResNeXt-d101-c32 (with SDPoint) to classify them correctly, in terms of top-5 accuracy. The ground-truth label names are shown below their corresponding images.}
	\label{fig:qualitative}
	\vspace*{-10pt}
\end{figure*}

\noindent\textbf{6.2.1\hspace{5pt}Ablation Study:} We study the effects of choice of SDPoint downsampling points and downsampling ratios on cost-adjustable inference performance. For this, we train a ResNeXt-d101-c32 with default SDPoint hyperparameters (downsampling points at the end of every residual block, downsampling ratios of \{0.5,0.75\}), as well as 2 baseline models with (i) downsampling points at the end of every other residual block dubbed \textit{\textbf{alternate}} (ii) downsampling ratio of just \{0.75\} dubbed \textit{\textbf{075}}. They are shown on Fig. \ref{fig:resnext}. Either removing the $0.5$ downsampling ratio or alternating blocks for downsampling gives worse results, due to reduced stochasticity (and regularization strengths).\\

\noindent\textbf{6.2.2\hspace{5pt}State-of-the-art Comparison:} We compare our models with SOTA ResNeXt-d101-c64 \cite{xie2017aggregated} and DenseNetBC-d264-g48 \cite{pleiss2017memory} models in Table \ref{table:imagenet}. SDPoint pushes the top-1 and top-5 validation errors of ResNeXt-d101-c32 down to 20.4\% and 5.3\% respectively, which are (previously) only attainable by SOTA models with roughly $ 2\times$ inference costs and parameter counts. We also display the results (and mean FLOPs) of Spatially Adaptive Computation Time (SACT) \cite{figurnov2017spatially} paired with PreResNet-d101, and compare it to a SDPoint instance of our PreResNet-d101 that achieves similar classification errors. SDPoint merely needs 69\% of FLOPs needed by SACT to achieve similar results. SACT saves computation by skipping layers (and network parameters) for certain locations in feature maps according to learned policy and inputs, while SDPoint downsamples feature maps to save computation (but makes full use of network parameters \& capacity during inference). We contend that in cost-accuracy trade-off for inference, reducing feature map spatial sizes is less harmful to accuracy than skipping network parameters/layers.\\

\noindent\textbf{6.2.3\hspace{5pt}Analysis:} We provide some analyses of ResNeXt-d101-c32 (trained with SDPoint on ImageNet) with regards to certain aspects of downsampling and SDPoint.\\
\vspace*{-6pt}

\noindent\textbf{Cost-dependent misclassifications:} We group ImageNet validation images (which are correctly classified with full inference cost) according to the minimum inference costs required to classify them correctly, and present some examples on Fig. \ref{fig:qualitative}. More difficult examples that require higher inference costs  (9.9, 16.0 GFLOPs) to be classified correctly, generally have size-dominant interfering objects/scenes (e.g., \texttt{hair dryer}, \texttt{cab}, \texttt{caldron}, \texttt{cock}, \texttt{tench}), in contrast to the easier examples (4.3 GFLOPs). Intuitively, pooling-based downsampling causes more information loss to smaller objects than to larger (size-dominant) objects, especially when it occurs at some early layer, where the semantic/context information is still relatively weak to distinguish objects of interest from interfering objects. So, for those difficult examples, it makes sense to preserve spatially informative object details longer in the CNN layer hierarchy, and downsample the feature maps only after they are semantically rich enough. \\

\noindent\textbf{Scale sensitivity:} Training CNNs with SDPoint involves stochastic downsampling of intermediate feature maps, which we hypothesize to be beneficial for scale sensitivity/invariance, as mentioned in Sect. \ref{regularization}. To validate this hypothesis, we vary the \textit{pre-cropping}\footnote{It is a standard practice \cite{he2016deep,he2016identity,xie2017aggregated,huang2017densely} to resize images to have a shorter side of 256 (\textbf{\textit{pre-cropping}} size) before doing $224\times224$ center-cropping.} sizes of ImageNet validation images in the range of $256,..., 352$ with step size of $16$, resulting in 7 \textit{pre-cropping} sizes. For every \textit{pre-cropping} size, $224\times224$ center image regions are cropped out for evaluation. The models involved are SDPoint-enabled ResNeXt-d101-c32, and the baseline without SDPoint. We compute the mean of all pairwise cosine similarities (a total of 21 pairs) resulted from the different \textit{pre-cropping} sizes, in terms of \textbf{ImageNet 1k-class probability scores}. This is done for entire ImageNet validation set. The pairwise cosine-similarity mean obtained for baseline model is \textbf{0.944}, while for the SDPoint-enabled model, it is \textbf{0.961}. A higher cosine similarity is a strong indicator of the model being less sensitive to scales. This demonstrates that SDPoint can indeed benefit CNNs, in terms of scale sensitivity. \\
\vspace*{-10pt}

\section{Conclusion}
We propose Stochastic Downsampling Point (SDPoint), a novel approach to train CNNs by downsampling intermediate feature maps. At no extra parameter and training costs, SDPoint facilitates effective cost-adjustable inference and greatly improves network regularization (thus accuracy performance). Through experiments, we additionally find out that SDPoint can help to identify more optimal (yet less costly) sub-networks (Sect. 6.1.2), sort input examples by various levels of classification difficulties (Fig. \ref{fig:qualitative}), and making CNNs less scale-sensitive (Sect. 6.2.3). 

{\small
\bibliographystyle{ieee}
\bibliography{egbib}
}

\end{document}